\documentclass[lettersize,journal]{IEEEtran}
\usepackage{amsmath,amsfonts,amssymb}
\usepackage{algorithmic}
\usepackage{algorithm}
\usepackage{graphicx}
\usepackage{textcomp}
\usepackage{url}
\usepackage{booktabs}
\usepackage{multirow}
\usepackage{multicol}
\usepackage{makecell}
\usepackage{enumitem}
\usepackage{xcolor}
\usepackage{colortbl}
\usepackage[numbers,sort&compress]{natbib}

\definecolor{mygray}{gray}{.9}
\definecolor{myblue}{rgb}{0.9, 0.9, 1}

\begin{document}

\title{AD-Copilot: A Vision-Language Assistant for Industrial Anomaly Detection via Visual In-context Comparison}

\author{Xi~Jiang,
        Yue~Guo,
        Jian~Li,
        Yong~Liu,
        Bin\mbox{-}Bin~Gao,
        Hanqiu~Deng,
        Jun~Liu,
        Heng~Zhao,
        Chengjie~Wang,
        and~Feng~Zheng~\IEEEmembership{Member,~IEEE}%
        \thanks{X. Jiang and F. Zheng are with the Department of Computer Science and Engineering, Southern University of Science and Technology, Shenzhen, China (e-mail: jiangx2020@mail.sustech.edu.cn, f.zheng@ieee.org).}%
        \thanks{Y. Guo is with Macau University of Science and Technology, Macau, China.}
        \thanks{J. Li, Y. Liu, B.-B. Gao, H. Deng, J. Liu, and C. Wang are with Tencent YouTu Lab, Shenzhen and Shanghai, China. J. Li is also with Nanjing University, Nanjing, China. C. Wang is also with Shanghai Jiao Tong University, Shanghai, China.}
        \thanks{H. Zhao is with the Centre for Frontier AI Research, A*STAR, Singapore.}
        \thanks{Code, model, and data will be released at \protect\url{https://github.com/jam-cc/AD-Copilot}.}
        }

\markboth{IEEE Transactions on Image Processing, Vol.~XX, No.~X, Month~20XX}%
{AD-Copilot: A Vision-Language Assistant for Industrial Anomaly Detection via Visual In-context Comparison}

\maketitle

\begin{abstract}
Multimodal Large Language Models (MLLMs) have achieved impressive success in natural visual understanding, yet they consistently underperform in industrial anomaly detection (IAD).
This is because MLLMs trained mostly on general web data differ significantly from industrial images. 
Moreover, they encode each image independently and can only compare images in the language space, making them insensitive to subtle visual differences that are key to IAD.
To tackle these issues, we present \textbf{AD-Copilot}, an interactive MLLM specialized for IAD via \textit{visual in-context comparison}.
We first design a novel data curation pipeline to mine inspection knowledge from sparsely labeled industrial images and generate precise samples for captioning, VQA, and defect localization, yielding a large-scale multimodal dataset \textbf{Chat-AD} rich in semantic signals for IAD. 
On this foundation, AD-Copilot incorporates a novel \textbf{Comparison Encoder} that employs cross-attention between paired image features to enhance multi-image fine-grained perception, and is trained with a \textbf{multi-stage strategy} that incorporates domain knowledge and gradually enhances IAD skills.
In addition, we introduce \textbf{MMAD-BBox}, an extended benchmark for anomaly localization with bounding-box-based evaluation.
The experiments show that AD-Copilot achieves 82.3\% accuracy on the MMAD benchmark, outperforming all other models without any data leakage.
In the MMAD-BBox test, it achieves a maximum improvement of $3.35\times$ over the baseline. AD-Copilot also exhibits excellent generalization of its performance gains across other specialized and general-purpose benchmarks.
Remarkably, AD-Copilot surpasses human expert-level performance on several IAD tasks, demonstrating its potential as a reliable assistant for real-world industrial inspection. 
Code and models are released in \url{https://github.com/jam-cc/AD-Copilot}.
\end{abstract}

\begin{IEEEkeywords}
Industrial Anomaly Detection, Multimodal Large Language Models, Visual In-context Comparison, Anomaly Localization, Vision-Language Assistant.
\end{IEEEkeywords}

%% ==================== FIGURE 1 (from main_cvpr.tex) ====================
\begin{figure*}[!t]
\centering
% \vspace{-0.5em}
\includegraphics[width=1.0\linewidth]{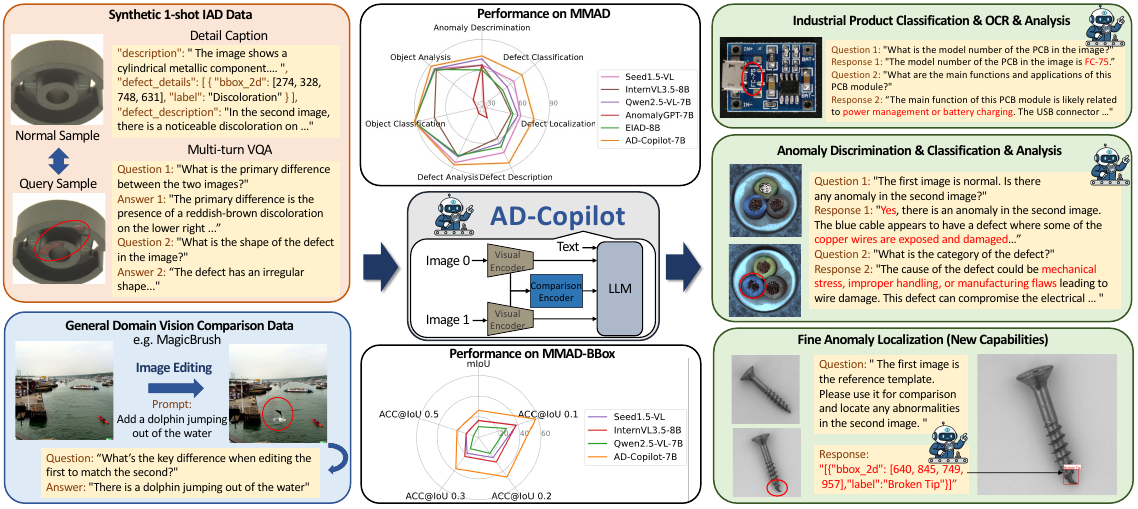}
\vspace{-0.5em}
\caption{AD-Copilot was trained on meticulously curated datasets, including high-quality synthetic IAD caption and VQA data, together with general-domain data aligned with IAD objectives. Experiment results show that AD-Copilot delivers strong performance across a range of IAD tasks. Case studies on downstream applications further highlight its practical utility and substantial potential in real-world manufacturing settings.}
\label{fig1}
% \vspace{-0.5em}
\end{figure*}

%% ==================== INTRODUCTION ====================
\section{Introduction}\label{sec:introduction}

Multimodal Large Language Models (MLLMs) have made remarkable progress in computer vision, demonstrating expert-level perception, judgment, and reasoning across a wide range of general-domain tasks~\citep{cheng2024domain, xu2025lingshu}.
Recent models such as Qwen2.5-VL~\citep{bai2025qwen2}, InternVL-3.5~\citep{wang2025internvl3}, and Seed-1.5-VL~\citep{guo2025seed1} further showcase strong visual grounding capabilities, opening new possibilities for fine-grained visual understanding.
Despite these advances, industrial anomaly detection (IAD)---a fundamental task in manufacturing quality control~\citep{jiang2022survey, wang2025softpatch+}---remains a critical frontier where MLLMs consistently fall short~\citep{jiangmmad, jin2024efficientMLLMsurvey}.
MLLMs hold great potential to transform industrial inspection by integrating perception, reasoning, and natural language communication within a single system.
This raises a central question: \textbf{\textit{how can we finetune general MLLMs to serve as reliable assistants for IAD?}}

There are two fundamental bottlenecks to be overcome.
First, there is a \textbf{scarcity of large-scale, high-quality industrial training data} tailored to the IAD setting.
Industrial images differ fundamentally from natural images~\citep{zhang2024ader}. They exhibit repetitive patterns, strict geometry, and subtle flaws that MLLMs easily overlook. Current MLLMs are trained mostly on general web data and rarely encounter industrial images, precise defect labels, or region-level anomaly cues~\citep{chen2025can}, leaving them without the domain knowledge needed to recognize and interpret industrial defects. 
Second, general MLLMs exhibit \textbf{limited capability in visual in-context comparison}, which is crucial for IAD.
In real-world quality inspection, human inspectors naturally compare products against standard templates to spot subtle deviations, such as minute scratches, color shifts, or structural irregularities, that become apparent only alongside a normal reference.
This comparison paradigm is so central to IAD that most benchmarks adopt a few-shot setting where normal templates accompany the query image~\citep{jiangmmad}.
However, existing MLLMs encode each image independently; any comparison occurs only at the language-reasoning stage, where the model can identify high-level semantic differences (\textit{e.g.}, ``a hole in a product'') but systematically fails to perceive subtle, pixel-level visual deviations~\citep{Li_2024_CVPR}.
Prior studies~\citep{jiangmmad} confirm that providing template images to existing MLLMs yields negligible improvement, precisely because fine-grained visual comparison is not performed.

% We term this the \textbf{semantic comparison bottleneck}.
% This bottleneck directly underlies the three widely recognized gaps in MLLM-based IAD---weak defect perception, limited generalization, and unreliable anomaly localization~\citep{zhang2024ader, chen2025can}.

Some recent work has made an effort in adapting general-purpose MLLMs with industrial multimodal data to build specialized models~\citep{fan2025manta,xu2025towards,li2025triad}. Pioneering models, such as AnomalyGPT~\citep{gu2024anomalygpt} and EIAD~\citep{zhang2025eiad}, directly translate MVTec-AD~\citep{bergmann2019mvtec} and VisA~\citep{zou2022spot} annotations into fixed-format Visual Question Answering (VQA) pairs for vision–language alignment and supervised finetuning. This reduces data collection cost, but the gains are limited by narrow data coverage, weak curation, and restricted linguistic diversity. Concurrently, advances in reasoning models, exemplified by OpenAI's o-series~\citep{jaech2024openai} and DeepSeek-R1~\citep{guo2025deepseek}, have advanced post-training techniques. Inspired by Reinforcement Learning with Verifiable Rewards (RLVR)~\citep{shao2024deepseekmath}, several studies bring this idea to IAD to boost reasoning and task performance~\citep{chao2025anomalyr1,zhao2025omniad,guan2025emit,liao2025ad}. Yet these Reinforcement Learning (RL) gains are capped when the base models lack strong IAD perception and localization; post-training cannot compensate for a weak foundation. 

In this work, we propose to address challenges through a holistic framework centered on visual in-context comparison for IAD, built upon three mutually reinforcing components:
\begin{enumerate}[leftmargin=*]

\item \textbf{Chat-AD Dataset (Data Perspective).}
We address the scarcity of high-quality IAD training data through large-scale collection from both public and proprietary sources, combined with a robust synthesis pipeline that generates contrastive descriptions, open-ended VQA pairs, verifiable questions, and reasoning data.
Critically, every sample pairs a query image with a normal template, and the generated text explicitly describes the visual differences between them---establishing \textit{comparison as the foundational data paradigm}.
The resulting dataset comprises over 620,000 samples across 327 industrial categories, filling a critical gap in the availability of comprehensive IAD training data for MLLMs.

\item \textbf{Comparison Encoder (Model Perspective).}
We design a novel architectural module that directly addresses the semantic comparison bottleneck.
Inspired by DETR~\citep{carion2020end}, the Comparison Encoder uses cross-attention between paired image features to generate a compact set of learnable comparison tokens that encode fine-grained visual differences.
Unlike approaches that modify image features in place, our design preserves the original image representations entirely, enabling plug-and-play integration with existing MLLM architectures without degrading their baseline capabilities.

\item \textbf{Multi-stage Training (Training Perspective).}
We develop a progressive training curriculum that builds comparison capability from the ground up.
Starting with general visual difference understanding on change detection datasets (Stage~0), the model learns to encode image differences into comparison tokens.
It then absorbs industrial knowledge through contrastive descriptions (Stage~1), refines its instruction-following abilities through multi-task dialogues (Stage~2), and finally enhances reasoning through reinforcement learning with task-specific verifiable rewards (Stage~3).

\end{enumerate}

\noindent In addition to our data and model contributions, we observe that existing benchmarks lack a fair evaluation of fine-grained anomaly localization for MLLMs.
Fine localization of anomalies is crucial in practice, as it helps distinguish defect types and provides actionable visualization for quality inspection~\citep{zhang2025eiad, li2026accurate}.
However, previous benchmarks~\citep{jiangmmad, xu2025towards} did not support this task, and localization was simplified as a multiple-choice task with descriptive phrase choices, which fails to reflect real industrial scenarios.
Meanwhile, recent MLLMs can already output bounding box coordinates as text~\citep{bai2025qwen2,wang2025internvl3,guo2025seed1}, creating a clear gap between model capability and benchmark support.
So, we propose \textbf{MMAD-BBox}, an extension to the MMAD benchmark that provides an impartial evaluation of MLLMs' fine localization performance in IAD.
To handle the irregular shapes and disconnected fragments typical of industrial defects, we further propose BBox-Mask IoU, a new metric that converts both predicted and ground-truth boxes into binary masks and computes IoU in the mask space, offering a fairer and more consistent measure than conventional box-matching IoU.

As shown in Fig.~\ref{fig1}, based on a general-purpose MLLM with our improved architecture, after multi-stage training on Chat-AD and several general-purpose visual comparison datasets, we obtain \textbf{AD-Copilot}, an interactive MLLM specialized for IAD. 
Benefiting from its enhanced fine-grained visual comparison capability, AD-Copilot excels at a wide spectrum of IAD tasks, including fine anomaly localization with bounding boxes.
We evaluate AD-Copilot on multiple benchmarks to validate the effectiveness of our visual in-context comparison framework.
AD-Copilot achieves 82.3\% accuracy on the MMAD~\citep{jiangmmad} benchmark, reaching the level of ordinary human performance and surpassing all proprietary and open-source models, including those trained with MMAD data.
On MMAD-BBox, it achieves a $3.35\times$ improvement in localization accuracy over the Qwen2.5-VL baseline. 
On the more specialized PCB inspection benchmark UniPCB~\citep{Sun2026UniPCBAU}, AD-Copilot also significantly outperforms other open-source models. On the general image comparison benchmark OmniDiff-Real\citep{liu2025omnidiff}, AD-Copilot also shows improvements over the baseline, demonstrating that the visual comparison gains from our framework generalize well. 
We name our model AD-Copilot to emphasize its role as a universal assistant, supporting human experts in understanding, analyzing, and localizing industrial anomalies through reliable, interactive, multimodal intelligence.

% \noindent Our main contributions can be summarized as follows:
% \begin{itemize}
% \item We construct \textbf{Chat-AD}, a large-scale, comparison-centric multimodal IAD dataset with over 620,000 high-quality samples and 2 million QA pairs across 327 industrial categories. 
% \item We propose the \textbf{Comparison Encoder}, a novel plug-and-play module that generates comparison tokens via cross-attention between paired images, enabling fine-grained visual comparison within existing MLLM architectures.
% \item We design a \textbf{multi-stage training curriculum} that progressively builds visual comparison capability, from general visual difference understanding to industrial anomaly detection and reasoning.
% \item We introduce \textbf{MMAD-BBox}, a new benchmark component that fills the gap in evaluating MLLMs’ fine localization capability for industrial anomalies.
% \item Extensive experiments validate that AD-Copilot demonstrates \textbf{state-of-the-art(SOTA)} performance across multiple industrial anomaly detection and localization tasks.
% \end{itemize}

The remainder of this paper is organized as follows.
Section~\ref{sec:related} reviews related work on IAD with MLLMs and visual comparison.
Section~\ref{sec:dataset} describes the Chat-AD dataset construction.
Section~\ref{sec:model} presents the model architecture with the Comparison Encoder.
Section~\ref{sec:training} details the multi-stage training strategy.
Section~\ref{sec:benchmark} introduces the MMAD-BBox benchmark.
Section~\ref{sec:experiments} reports comprehensive experimental results and analyses.
Section~\ref{sec:conclusion} concludes the paper.

%% ==================== RELATED WORK ====================
\section{Related Work}\label{sec:related}

\subsection{Industrial Anomaly Detection with Foundation Models}

Traditional IAD methods are usually trained from sets of normal images~\citep{jiang2022softpatch, deng2022anomaly, batzner2024efficientad}.
As a result, they only work on product categories that have been seen during training. 
Even in recent studies on multi-class IAD~\citep{ding2025focuspatch, you2022unified, jiang2024toward}, models still fail to benefit from larger datasets in generalizing to new, unseen categories.
Multimodal foundation models have introduced new possibilities to IAD, enabling it to learn the general pattern of anomaly detection from limited data and perform few-shot or even zero-shot detection on unseen categories~\citep{wang2025normal, dai2024seas, li2024musc}.
They can be roughly divided into two groups.
The first group is based on CLIP series models, which establish a unified vision-language space for industrial anomaly discrimination~\citep{jeong2023winclip, cao2024adaclip, zhou2023anomalyclip, Li_2024_CVPR, sadikaj2025multiads}. 
The second group focuses on MLLMs, which distinguish anomalies with language models~\citep{cao2023towards, xu2025customizing, zhu2024llms, zhang2024gpt}. Since these MLLMs generate language outputs, they can also address other quality assurance problems, including defect classification, description, and analysis. 

After the release of benchmark MMAD~\citep{jiangmmad}, numerous methods have been proposed to enhance MLLMs for IAD tasks.
Some works~\citep{mokhtar2025detect, chen2025can} explore training-free prompt engineering to guide models with additional visual or textual information.
Others~\citep{zhang2025eiad, xu2025towards, li2025triad, li2025iad} build large-scale IAD training datasets and train MLLMs on them. 
However, most of these datasets are presented only in a simple VQA format, which lacks linguistic and task diversity.   
A growing number of studies~\citep{zeng2025lr, chao2025anomalyr1, zhao2025omniad, guan2025emit, liao2025ad} focus on RL finetuning to rapidly improve MLLM performance in IAD. 
These RL methods have achieved large performance gains on MMAD, but their generalization ability remains uncertain, as they often perform domain adaptation using partial MMAD data, which introduces potential data leakage.

\subsection{Visual Comparison and Change Detection}

Change detection and change captioning have established a rich foundation for visual comparison between image pairs.
In remote sensing, ChangeFormer~\citep{bandara2022changeformer} demonstrates that transformer-based Siamese networks can capture long-range inter-image dependencies for accurate bi-temporal change detection, outperforming CNN-based counterparts.
For natural-language descriptions of differences, CLEVR-Change~\citep{park2019robust} and Spot-the-diff~\citep{jhamtani2018learning} establish the change captioning task, which requires fine-grained visual comparison and joint localization; Chg2Cap~\citep{chang2023chg2cap} further bridges change detection and captioning in remote sensing via a hierarchical attentive encoder.
Image editing datasets such as MagicBrush~\citep{Zhang2023MagicBrush} extend this to instructional editing scenarios, providing paired before-and-after images with textual change descriptions.

More recently, the rise of MLLMs has stimulated new approaches to general-domain image difference captioning (IDC).
OneDiff~\citep{hu2024onediff} builds a generalist IDC model via a siamese encoder with a Visual Delta Module for multi-scale difference extraction, while Img-Diff~\citep{jiao2024imgdiff} synthesizes contrastive paired data to improve MLLMs' fine-grained visual comparison abilities.
OmniDiff~\citep{liu2025omnidiff} provides a comprehensive IDC benchmark across 324 diverse real-world and synthetic scenarios.
However, these works mainly focus on general-purpose tasks. In IAD, by contrast, the differences to be detected are often less semantic and more subtle~\citep{liao2024coft, zhang2025multi}, and the decision also requires inherent knowledge. Therefore, we design a new plug-and-play comparison module that compresses comparison information into a small set of tokens, enabling the improvement in IAD task based on its original perception and reasoning capabilities. 

%% ==================== CHAT-AD DATASET ====================
\section{Chat-AD Dataset}\label{sec:dataset}

\begin{figure*}[htb]
\centering
% \vspace{-0.8em}
\includegraphics[width=1.0\linewidth]{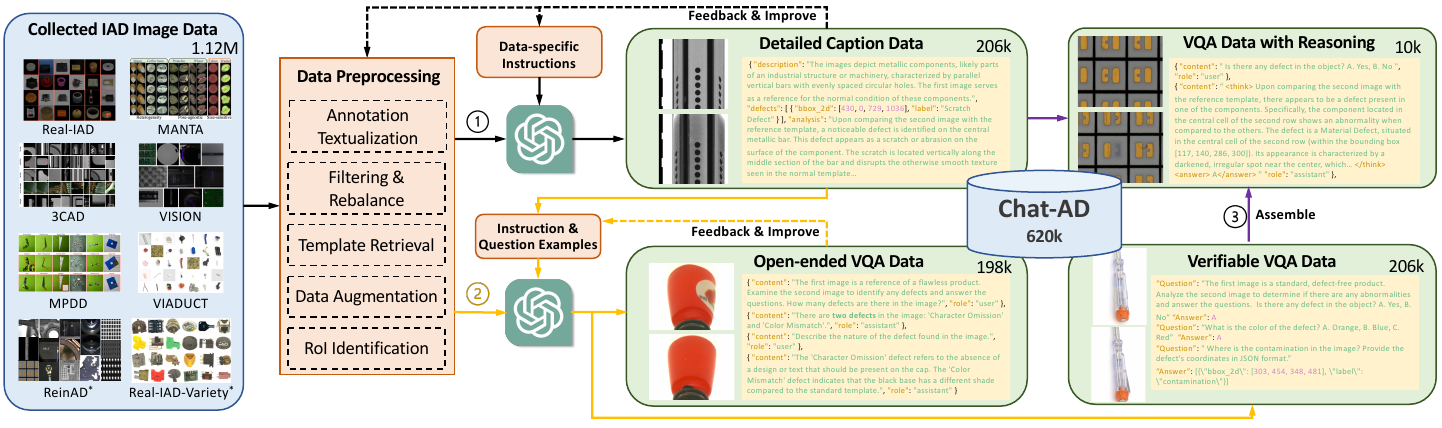} 
\vspace{-0.8em}
\caption{The pipeline for Chat-AD synthesis. Each query image is paired with a normal template image, and the generated text explicitly describes visual differences between them, establishing comparison as the foundational data paradigm. $^{\ast}$~Currently private, upcoming public dataset.}\label{chat-ad-fig}
\vspace{-0.8em}
\end{figure*}

To address the scarcity of detailed textual annotations in IAD datasets, we construct four types of image-text data to help MLLMs learn industrial anomaly knowledge more effectively.
We comprehensively collect large-scale multi-category IAD datasets, including VIADUCT~\citep{lehr2024ad3} with the richest object categories, Real-IAD~\citep{wang2024real} captured from multi-view production lines, MANTA~\citep{fan2025manta} containing diverse tiny objects, 3CAD~\citep{yang20253cad} with various 3C products, and MPDD~\citep{jezek2021deep} covering diverse object poses.
Additionally, we include two private datasets, Real-IAD-Variety and ReinAD, to further enhance the diversity of objects and anomalies. Notably, these datasets share no overlapping categories with those datasets~\citep{bergmann2019mvtec, bergmann2022beyond, zou2022spot, zhang2024pku} in MMAD.

As shown in Fig.~\ref{chat-ad-fig}, the final collection comprises approximately 1.12 million industrial images.
Since most datasets consist of repetitive normal samples, after applying image quality filtering, we rebalance the ratio of abnormal to normal samples for each dataset.

\paragraph{Query-Template Pairing: The Foundation of Visual In-context Comparison}
To establish visual in-context comparison as the core data paradigm, each query image is paired with a similar template image retrieved from the normal pool.
This design directly mirrors real-world quality inspection workflows, where human inspectors compare products against standard references.
By consistently providing paired images during training, we encourage the model to learn a general paradigm for visual anomaly detection based on comparison, rather than relying on memorized defect patterns.
Both query and template images undergo the same data augmentation, mainly cropping to enhance visual diversity.

\paragraph{Contrastive Description Generation}
We integrate existing annotations into metadata, including the text of object and defect categories, as well as visual annotations, where anomalies are highlighted by contour lines and bounding boxes to guide the following text generation.
We first produce long-form captions based on the preprocessed metadata. 
Critically, these captions are not simple image descriptions---they are \textit{contrastive descriptions} that explicitly emphasize the differences between the anomalous query and its normal template.
This contrastive framing teaches the model to articulate what it should look for when comparing two images, directly reinforcing the visual in-context comparison paradigm.
Iterative refinement and verification of outputs ensure accuracy and consistency. GPT-4o~\citep{achiam2023gpt} and Qwen2.5-VL-72B~\citep{bai2025qwen2} serve as our core data generators, with GPT-4o mainly used for early-stage pipeline validation and refinement, and Qwen2.5-VL-72B responsible for large-scale caption generation.

\paragraph{Multi-task Dialogue and Verifiable QA Data}
After obtaining high-quality image-caption pairs, we produce IAD dialogue data, including open-ended multi-turn conversations and verifiable question-answer pairs. 
The verifiable tasks, such as multiple-choice questions and fine anomaly localization tasks, enable the calculation of correctness scores, facilitating RLVR. 
Since the caption data already provides rich reasoning information, we combine verifiable QA data with caption samples to create IAD thinking data, which serves as examples for performing reasoning.

Finally, we construct the largest multimodal IAD dataset, \textbf{Chat-AD}, comprising over 620,000 high-quality samples and 2 million QA pairs across 327 industrial classes, filling a critical gap in the availability of comprehensive IAD training data for MLLMs.

%% ==================== MODEL ARCHITECTURE ====================
\section{Model Architecture}\label{sec:model}

\begin{figure*}[htb]
  \centering
  % \vspace{-0.8em}
\includegraphics[width=0.9\linewidth]{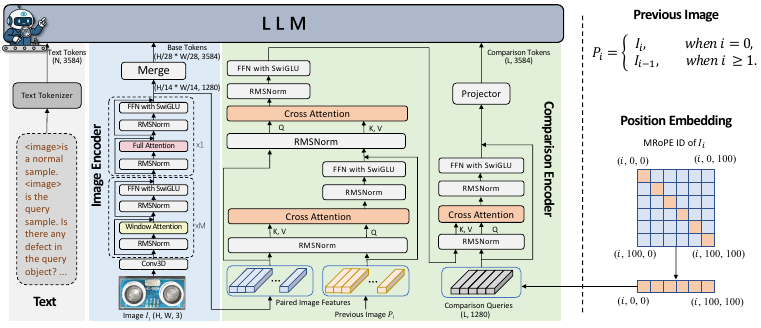} 
\vspace{-0.8em}
\caption{The model architecture of AD-Copilot. The Comparison Encoder performs cross-attention between paired image features to generate comparison tokens, enabling visual in-context comparison at the encoding stage.}\label{Structure}
\vspace{-0.8em}
\end{figure*}

\subsection{Motivation: The Semantic Comparison Bottleneck}

Existing MLLMs usually follow a two-stage structure, which extracts vision features aligned to the text space and then performs language-guided reasoning. This causes two key issues. First, the feature extraction process may lead to information loss, making the model insensitive to fine-grained cues. Second, each image is encoded independently, lacking inter-image interaction. As a result, these models can only compare images semantically rather than visually. In IAD, this means they can observe defects with clear semantics, such as holes, but often miss subtle anomalies like scratches. Consequently, previous models cannot achieve a significant performance gain from the template image~\citep{jiangmmad}. Comparing with templates to find anomalies is important in practice, while the default 1-shot setting provides a normal template image for each query image. 

\subsection{The Comparison Encoder}

To address this bottleneck, we introduce a \textbf{Comparison Encoder} that performs visual-level comparison between images at the encoding stage, as illustrated in Fig.~\ref{Structure}.
Building on the Qwen2.5-VL architecture, we add an inter-image interaction module at the feature level.

\paragraph{Architecture Design}
Inspired by DETR~\citep{carion2020end}, we propose a cross-attention-based module that encodes comparison information between each pair of images into a fixed set of 100 comparison tokens.
Before feature merging, fine-grained image features from the vision encoder are extracted and cross-attended to those from the paired image.
A set of learnable comparison queries then compresses the cross-attention results into a fixed-length representation, adding minimal computational overhead.

\paragraph{Design Rationale}
A key design choice, distinct from other IDC models, is to generate \textit{additional} comparison tokens rather than modifying the original image features in place.
This DETR-style approach has two important advantages:
(1)~it preserves the original image representations entirely, ensuring that the base model's capabilities are not degraded;
and (2)~it produces compact comparison tokens, enabling the LLM to further reason over them for inspection decisions. 

\paragraph{Structure Details}
We enable the model to adapt to a variable number of input images by pairing each image only with its immediately preceding counterpart (with self-pairing for the first image), thus achieving streaming extraction of contextual comparison information.
For the comparison tokens, we reuse the positional encoding along the image diagonal to capture a wider spatial range.
The number of comparison tokens L is determined by the diagonal length of the default image grid ($\sqrt{10000} = 100$), which provides sufficient spatial coverage for encoding fine-grained visual differences.
The added comparison tokens, limited in number, do not affect the original model’s pre-training performance, and after training on image comparison tasks, they effectively encode fine-grained inter-image differences within a few tokens, offering complementary cues for downstream reasoning.

%% ==================== MODEL TRAINING ====================
\section{Model Training}\label{sec:training}

AD-Copilot consists of a vision encoder, a comparison encoder, and an LLM.
Following previous work~\citep{xu2025lingshu, zhao2025omniad}, we initialize the vision encoder and LLM with Qwen2.5-VL-7B-Instruct~\citep{bai2025qwen2}, which provides strong multimodal understanding and instruction-following abilities.
The comparison encoder is initialized with a Gaussian distribution, providing a smooth start for learning fine-grained inter-image differences.

\subsection{Training Recipe: Progressive Visual Comparison Learning}\label{subsec:recipe}

Inspired by recent LLM and MLLM training paradigms~\citep{xu2025lingshu, li2023llava, olmo20242}, we propose a multi-stage curriculum that progressively builds the model's visual comparison capability and adapts it to IAD scenarios.
This framework follows a ``perception-to-reasoning'' progression, comprising four sequential stages.
Each stage incrementally enhances the model's ability to compare, perceive, and reason about industrial anomalies.

The overall training pipeline is summarized in Tab.~\ref{tab:data_mixture_english}.

\begin{table*}[htb]
\centering
\caption{Overview of data mixture for each training stage. All datasets have been curated through our data pipeline (Sec.~\ref{sec:dataset}).}
\label{tab:data_mixture_english}
\resizebox{\linewidth}{!}{
\begin{tabular}{>{\raggedright}p{5cm} >{\raggedright}p{14cm} r}
\toprule
\textbf{Stage} & \textbf{Training Data Composition} & \textbf{Amount} \\
\midrule
Stage 0: Learning to Compare \newline Full Finetuning: Comparison Encoder & \textbf{1. Image Contrastive Description (30k):} \newline CLEVR-Change(9k), MagicBrush(9k), Spot-the-diff(12k) & $\sim$30k \\
\rowcolor{gray!15}
Stage 1: Industrial Comparison \newline LoRA Finetuning: Comparison Encoder + LLM & \textbf{1. Industrial Image Contrastive Description (166k):} \newline Caption\_3CAD(16.1k), Caption\_MANTA(48.5k), Caption\_Real\_IAD(64k), Caption\_Real\_IAD\_Variety(26.2k), Caption\_VISION(1.6k), Caption\_MPDD(0.5k), Caption\_VIADUCT(5.2k) \newline \textbf{2. General Image Description (22k):} \newline PixMo-cap(22k) \newline \textbf{3. Image Contrastive Description (30k):} \newline CLEVR-Change(8.9k), MagicBrush(8.8k), Spot-the-diff(12.5k) & $\sim$218k \\
Stage 2: Multi-task Comparison \newline LoRA Finetuning: Image Encoder + Comparison Encoder + LLM & 
\textbf{1. Industrial Multi-turn Dialogue (198k):} \newline Chat\_3CAD(16.1k), Chat\_MANTA(48.5k), Chat\_Real\_IAD(59.6k), Chat\_Real\_IAD\_Variety(26.1k), Chat\_VISION(1.6k), Chat\_MPDD(0.5k), Chat\_VIADUCT(5.2k), Chat\_reinad(40k) \newline \textbf{2. General Image Dialogue (50k):} \newline M4\_Instruct(25k), LLaVA\_V1.5\_mix665k(25k) \newline \textbf{3. Pure Text (50k):} \newline OLMo-2-SFT(50k) & $\sim$298k \\
\rowcolor{gray!15}
Stage 3: Comparison-aware Reasoning \newline LoRA Finetuning: LLM & 
\textbf{1. Industrial Multiple-choice \& Localization:} \newline Chat\_3CAD\_RL(16.1k), Chat\_MANTA\_RL(48.5k), Chat\_Real\_IAD\_RL(59.6k), Chat\_Real\_IAD\_Variety\_RL(26.1k), Chat\_VISION\_RL(1.6k), Chat\_MPDD\_RL(0.5k), Chat\_VIADUCT\_RL(5.2k), Chat\_reinad\_RL(40k) & $\sim$206k \\
\bottomrule
\end{tabular}
}
\end{table*}

\textbf{Stage 0: Learning to Compare (Comparison Encoder Pretrain).}
This foundational stage teaches the comparison encoder to perceive visual differences between paired images.
The training focuses on contrastive and fine-grained change understanding using general datasets such as CLEVR-Change~\citep{park2019robust}, MagicBrush~\citep{Zhang2023MagicBrush}, and Spot-the-diff~\citep{jhamtani2018learning}.
We unify them into a two-image comparison task with instruction-style samples, enabling the comparison encoder to develop a general capability for representing image differences.
Only the comparison encoder is fully finetuned; the LLM is frozen to preserve its language generation capability.

\textbf{Stage 1: Learning Industrial Comparison (IAD Knowledge Alignment).}
This stage transfers the general comparison capability to the industrial domain.
Training is conducted on IAD image-text pairs where the captions not only describe the inspected object but also \textit{explicitly emphasize the differences between its abnormal and normal states}, specifying the detailed locations of defects.
This contrastive description format directly reinforces the visual in-context comparison paradigm, enabling AD-Copilot to connect industrial visual semantics with linguistic representations.

\textbf{Stage 2: Multi-task Comparison (IAD Instruction Tuning).}
At this stage, we unlock all modules and perform large-scale, end-to-end optimization using an instruction tuning corpus.
The instructions encompass a range of tasks, including anomaly judgment, defect classification, abductive reasoning about causes, defect localization, object analysis, and product counting.
All samples are presented as multi-turn dialogues to encourage deeper analytical reasoning.
To preserve general instruction-following capabilities, we supplement filtered general-domain resources such as OLMo-2-SFT~\citep{olmo20242}, PixMo-cap~\citep{deitke2025molmo}, and M4\_instruction~\citep{li2024llava}.
After this stage, the model's ability to perform diverse IAD tasks through visual comparison is substantially improved; we refer to it as \textit{AD-Copilot}.

\textbf{Stage 3: Comparison-aware Reasoning (IAD-oriented RL).}
Supervised finetuning may lead to overfitting~\citep{chu2025sft}, and its cross-entropy loss is suboptimal for continuous tasks such as fine localization~\citep{jiang2025detect}.
To address these limitations, we adopt GRPO-based reinforcement learning~\citep{guo2025deepseek} with verifiable rewards on IAD verifiable VQA data, which includes both multiple-choice (MC) and anomaly localization (LOC) tasks.
The overall reward function is defined as:
\begin{equation}
R(x,y) = \lambda R_{\text{fmt}}(y) +
\begin{cases}
\mathbf{1}[C_{\text{pred}} = C_{\text{gt}}], & x\!\in\!\text{MC},\\[4pt]
\mathbf{IoU}(B_{\text{pred}}, M_{\text{gt}}), & x\!\in\!\text{LOC},
\end{cases}
\end{equation}
where $R_{\text{fmt}}(y)$ encourages the model to output reasoning in \texttt{<think></think>} and the answer in \texttt{<answer></answer>}.
By incorporating localization-aware rewards, our RL stage provides finer supervision beyond discrete correctness signals used in prior approaches~\citep{zeng2025lr, chao2025anomalyr1, guan2025emit}.
After RL, the model improves both accuracy and interpretability, producing explicit reasoning traces. We denote this final checkpoint as \textit{AD-Copilot-Thinking}.

\subsection{Implementation Details}

The IAD tasks require high-resolution images for fine-grained perception. We set the maximum sequence length to 20k tokens and allow image resolutions up to 2800$\times$2800, resulting in high GPU memory consumption. Therefore, we apply LoRA finetuning (rank=64, $\alpha$=128) with DeepSpeed in all stages except Stage~0. In Stage~0, since the data do not require high resolution, we adopt full-parameter finetuning to ensure the comparison encoder is sufficiently trained. We use the LlamaFactory framework~\citep{zheng2024llamafactory} in SFT stages and the PR1 framework~\citep{yu2025perceptionr} in the RL stage. For all stages, the per-device batch size is set to 1, with a gradient accumulation of 16 for SFT stages and 2 for RL training. We train each stage for 2 epochs using a cosine learning rate scheduler with an initial learning rate of $5\times10^{-5}$. In the RL stage, we apply early stopping, as it converges much faster than SFT. The total training cost across all four stages is approximately 2,400 GPU hours on a single node.

%% ==================== MMAD-BBOX BENCHMARK ====================

\begin{figure}[htb]
  \centering
\includegraphics[width=0.9\linewidth]{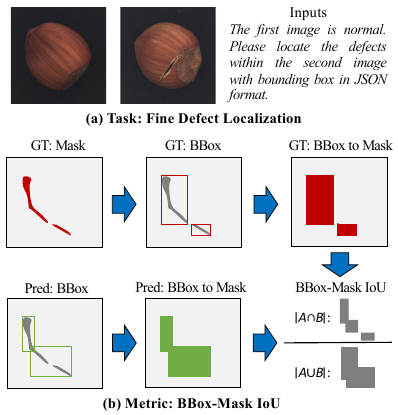} 
\vspace{-0.8em}
\caption{Task and metric of MMAD-BBox.}\label{iou}
\vspace{-0.8em}
\end{figure}

\section{MMAD-BBox Benchmark}\label{sec:benchmark}
Fine localization of anomalies is crucial in practice, as it helps distinguish defects and provides actionable visualization~\citep{zhang2025eiad}.
However, previous MLLMs lack the grounding ability, and earlier benchmarks~\citep{jiangmmad, xu2025towards} did not support this task.
Recent MLLMs can already output bounding box coordinates as text, which highlights the gap between model capability and benchmark support.
In the MMAD benchmark~\citep{jiangmmad}, localization was simplified as a multiple-choice task with descriptive phrase choices, which fails to reflect real industrial scenarios. 

To bridge this gap, we extend MMAD to support fine-grained anomaly localization and propose a new benchmark, \textbf{MMAD-BBox}, which evaluates bounding box predictions directly.
It utilizes the same image sources and default settings as MMAD, but does not rely on synthesized text data, thereby making the evaluation more realistic and reliable.

\begin{table*}[htb]
\centering
\setlength\tabcolsep{3pt}
\caption{Performance comparison of both proprietary and open-source MLLMs in MMAD with the standard 1-shot setting. $^{\dagger}$~Thinking model. $^{\ast}$~Models trained with MMAD data and used domain knowledge.} 
\label{mmad-performance-comparison}
% \resizebox{\linewidth}{!}
{
\begin{tabular}{c|c|c|cccc|cc|c}
\toprule\toprule
                         &                         & Anomaly        & \multicolumn{4}{c|}{Defect}                             & \multicolumn{2}{c|}{Object} &                           \\\cmidrule(r){3-9}
\multirow{-2}{*}{Model} & \multirow{-2}{*}{Scale} & Discrimination & Classification & Localization & Description & Analysis & Classification  & Analysis & \multirow{-2}{*}{Average} \\\midrule
\rowcolor{myblue} 
Human (expert)          & -                       & 95.24           & 75.00           & 92.31         & 83.33     & 94.20        & 86.11            & 80.37     & 86.65                     \\
\rowcolor{myblue} 
Human (ordinary)        & -                       & 86.90          & 66.25          & 85.58        & 71.25    & 81.52       & 89.58           & 69.72    & 78.69  
\\ \midrule
 Gemini-1.5-pro           & -    & 68.63          & 60.12          & 58.56        & 70.38       & 82.46    & 89.20           & 82.25    & 73.09                     \\
 GPT-4o                   & -    & 68.63          & 65.80          & 55.62        & 73.21       & 83.41    & \textbf{94.98}           & 82.80       & 74.92                     \\
 Seed1.5-VL               & -    & 65.30 & 64.32 & 63.35 & 73.80 & 84.38 & 91.65 & 83.67 & 75.21 \\
 Seed1.5-VL$^{\dagger}$   & -    & 55.86 & 62.36 & 63.03 & 72.36 & 83.69 & \underline{94.46} & 84.83 & 73.80 \\
\midrule
InternVL2                & 8B   & 59.97  & 43.85  & 47.91  & 57.60  & 78.10  & 74.18  & 80.37  & 63.14 \\
InternVL2                & 76B      & 68.25          & 54.22          & 56.66        & 66.30       & 80.47    & 86.40           & 82.92    & 70.75                    \\
Qwen2.5-VL               & 7B       & 71.10  & 56.02  & 60.69  & 64.13  & 78.26  & 91.49  & 83.67  & 72.19 \\
Qwen2.5-VL               & 72B      & 72.66  & 62.31  & 67.16  & 73.56  & 81.95  & 94.30  & 86.78  & 76.96 \\
InternVL3.5 & 8B & 64.93 &	48.33 &	53.83 &	58.65 &	78.00 &	71.97 &	81.24 &	65.28 \\
Qwen3-VL                 & 8B       & 68.71  & 63.27  & 61.01  & 68.18  & 79.99  & 92.27  & 83.96  & 73.91 \\
Qwen3-VL$^{\dagger}$     & 8B       & 56.29  & 62.17  & 60.93  & 75.34  & 84.58  & 93.59  & \underline{84.26}  & 73.88 \\
\midrule 
AnomalyGPT (2024,AAAI)              & 7B       & 65.57  & 27.49  & 27.97  & 36.86  & 32.11  & 29.84  & 35.82  & 36.52 \\
EIAD  (2025,ICME)                  & 8B       & 60.50  & 50.70          & 55.60        & 67.80       & 76.90   & 91.80     & 82.70       & 69.40 \\
Triad (2025,ICCV)                  & 7B       &    -    &    -    &   -     &    -    &    -    &     -   &     -   & 70.92 \\
\textcolor{gray}{AnomalyR1$^{\dagger\ast}$ (2025)}   & \textcolor{gray}{3B}   & \textcolor{gray}{60.62}  & \textcolor{gray}{63.56}  & \textcolor{gray}{70.14}  & \textcolor{gray}{80.47}  & \textcolor{gray}{85.28}  & \textcolor{gray}{92.48}  & \textcolor{gray}{86.15}  & \textcolor{gray}{76.96} \\
\textcolor{gray}{OmniAD$^{\dagger\ast}$ (2025)}      & \textcolor{gray}{7B}   & \textcolor{gray}{68.80}  & \textcolor{gray}{78.80}  & \textcolor{gray}{75.50}  & \textcolor{gray}{67.20}  & \textcolor{gray}{86.40}  & \textcolor{gray}{96.00}  & \textcolor{gray}{86.40}  & \textcolor{gray}{79.90} \\
\textcolor{gray}{EMIT$^{\dagger\ast}$ (2025)}        & \textcolor{gray}{8B}   & \textcolor{gray}{73.87}  & \textcolor{gray}{80.85}  & \textcolor{gray}{76.39}  & \textcolor{gray}{83.00}  & \textcolor{gray}{85.92}  & \textcolor{gray}{90.26}  & \textcolor{gray}{83.37}  & \textcolor{gray}{81.95} \\
AD-Copilot              & 7B       & \underline{73.64}  & \underline{67.89}  & \underline{64.08}  & \underline{80.60}  & \underline{85.91}  & 91.06  & \textbf{87.78}  & \underline{78.71} \\
AD-Copilot$^{\dagger}$  & 7B       & \textbf{73.95}  & \textbf{74.29}  & \textbf{76.40}  & \textbf{84.92}  & \textbf{86.93}  & 91.86  & \underline{87.67}  & \textbf{82.29} \\
\bottomrule
\bottomrule
\end{tabular}
}
\end{table*}

\paragraph{BBox-Mask IoU}
Conventional box-matching IoU metrics can be unreliable, as industrial defects often have irregular shapes, disconnected fragments, or ambiguous instance boundaries.
To provide a fairer and more consistent evaluation, we propose a new metric called \textbf{BBox-Mask IoU}.
As illustrated in Fig.~\ref{iou}, we convert both the predicted boxes and the ground-truth boxes into binary masks and compute IoU in the mask space.
This alleviates bias caused by differing output granularities and offers a smooth, fine-grained measure of localization accuracy.
In addition to the mean IoU, we report accuracy under IoU thresholds of 0.5, 0.3, 0.2, and 0.1, where relatively low thresholds are adopted as anomaly localization is inherently more challenging than standard object detection.

%% ==================== EXPERIMENTS ====================
\section{Experiments}\label{sec:experiments}

\subsection{Experimental Setup}

We evaluate AD-Copilot on four benchmarks: the MMAD and the proposed MMAD-BBox benchmarks for industrial anomaly detection, UniPCB~\citep{Sun2026UniPCBAU} for PCB-domain generalization, and OmniDiff-Real~\citep{liu2025omnidiff} for general visual comparison beyond the industrial domain.
For MMAD and MMAD-BBox, we use the default 1-shot setting without domain knowledge; for MMAD-BBox, we carefully adapt prompts for different MLLMs and allow them to output bounding boxes in different formats.

We compare AD-Copilot against three categories of state-of-the-art models:
\begin{itemize}[leftmargin=*]
    \item \textbf{Proprietary Models:} GPT-4o~\citep{achiam2023gpt}, Gemini~1.5~Pro~\citep{reid2024gemini}, and Seed-1.5-VL~\citep{guo2025seed1}.  
    \item \textbf{General-purpose MLLMs:} Qwen2.5-VL~\citep{bai2025qwen2}, Qwen3-VL~\citep{team2025qwen3}, InternVL2~\citep{chen2023internvl}, and InternVL3.5~\citep{wang2025internvl3}.
    \item \textbf{IAD-specific MLLMs:} AnomalyGPT~\citep{gu2024anomalygpt}, EIAD~\citep{zhang2025eiad}, TriAD~\citep{li2025triad}, AnomalyR1~\citep{chao2025anomalyr1}, OmniAD~\citep{zhao2025omniad}, and EMIT~\citep{guan2025emit}.  
\end{itemize}
 
\noindent Note that AnomalyR1, OmniAD, and EMIT are trained using a portion of the MMAD test data and are therefore not directly comparable to our model. We mark these methods with gray text in result tables to indicate this distinction.

\subsection{Performance Comparison}

\paragraph{MMAD benchmark}
Tab.~\ref{mmad-performance-comparison} presents a detailed comparison between AD-Copilot and a diverse set of both proprietary and open-source MLLMs across seven MMAD sub-tasks.  
Among general-purpose models, the Qwen2.5-VL-72B achieves the highest average accuracy of 76.96.  
In contrast, existing IAD-specific models, which are mostly developed upon smaller and older foundation models, reach at most 70.92.  
Our proposed AD-Copilot significantly advances this line of work: the 7B \textit{thinking} model achieves the best overall accuracy of 82.29, while the \textit{non-thinking} model ranks second with 78.71---representing a 10.1-point improvement over the Qwen2.5-VL-7B baseline.  
This performance not only surpasses all open-source and proprietary models but also exceeds the average accuracy of ordinary human evaluators.  
Notably, it outperforms several IAD-specific models that were trained with MMAD data and utilize domain knowledge, highlighting the robustness and generalization of our framework.

Specifically, AD-Copilot reaches expert-level accuracy in \textit{defect classification} and even surpasses human experts in \textit{defect description}, \textit{object classification}, and \textit{object analysis}.  
However, a noticeable gap remains in \textit{anomaly discrimination} and \textit{defect localization}, two key capabilities for real-world quality inspection.  
These results indicate that while current MLLMs cannot yet fully replace human inspectors, AD-Copilot has reached a level where it can serve as a reliable and insightful assistant in industrial inspection workflows.

\begin{table}[htb]
\centering
\setlength\tabcolsep{5pt}
\caption{Comparison on the MMAD-BBox benchmark (1-shot).}
\label{localization-comparison}
\resizebox{\linewidth}{!}{
\begin{tabular}{c|c|c|cccc}
\toprule\toprule
& &  & \multicolumn{4}{c}{ACC@IoU(\%)} \\\cmidrule(r){4-7}
\multirow{-2}{*}{Model} & \multirow{-2}{*}{Scale} & \multirow{-2}{*}{\makecell[c]{mIoU\\(\%)}} & 0.1 & 0.2 & 0.3 & 0.5 \\
\midrule
Qwen2.5-VL & 7B & 10.47 & 28.39 & 18.83 & 13.39 & 5.15 \\
Qwen3-VL & 8B & 9.30 & 25.14 & 17.39 & 12.30 & 5.74 \\
InternVL3.5 & 8B & 16.34 & 38.15 & 29.26 & 22.63 & 12.54 \\
Seed1.5-VL & - & 16.58 & 38.62 & 23.90 & 18.81 & 12.50 \\
Qwen2.5-VL(Finetuned) & 7B & 23.30 & 54.08 & 41.44 & 32.39 & 17.35 \\
AD-Copilot & 7B & \underline{24.46} & \textbf{55.66} & \underline{43.77} & \underline{34.78} & \underline{19.42} \\
AD-Copilot-Thinking & 7B & \textbf{25.30} & \underline{55.22} & \textbf{44.76} & \textbf{35.88} & \textbf{21.00} \\
\bottomrule\bottomrule
\end{tabular}
}
\end{table}
\paragraph{MMAD-BBox benchmark}
The frontier MLLMs such as Qwen3-VL, InternVL3.5, and Seed1.5-VL have demonstrated strong single-image grounding performance. However, as shown in Tab.~\ref{localization-comparison}, when evaluated on anomaly localization, their performance drops drastically. Seed1.5-VL performs best among general-purpose models, but with an mIoU of only 16.58. Even with a loose IoU threshold of 0.1, the accuracy reaches only 38.6\%. This poor performance is likely due to the domain shift of anomaly images and the need for multi-image perception, which poses greater challenges than standard grounding tasks.
In contrast, AD-Copilot achieves a clear lead, with an mIoU of 25.30 and an IoU@0.5 accuracy of 21\%, representing a $3.35\times$ improvement over the baseline (5.15\%).  
We also finetuned Qwen2.5-VL-7B using the same training strategy as AD-Copilot but without the Comparison Encoder, which led to noticeable gains yet remained slightly behind AD-Copilot, further demonstrating the effectiveness of both the training strategy and the Comparison Encoder.

\paragraph{UniPCB benchmark}
As shown in Tab.~\ref{tab:unipcb}, AD-Copilot significantly outperforms all evaluated released MLLMs on UniPCB~\citep{Sun2026UniPCBAU}, the newest multi-task benchmark for PCB inspection, including the IAD-specific EMIT model, demonstrating strong transferability to unseen industrial domains. However, AD-Copilot shows a slight drop in F1-score for PCB localization, which may stem from its lack of prior knowledge of PCB products and defects in Chat-AD, leading to suboptimal localization accuracy. 
\begin{table*}[htb]
\centering
% \caption{Performance comparison on the UniPCB and OmniDiff-Real benchmarks.}
% \label{tab:combined}
\begin{minipage}[ht]{0.55\textwidth}
\centering
\caption{Performance comparison on the UniPCB Benchmark.}
\label{tab:unipcb}
\setlength\tabcolsep{3pt}
\resizebox{\linewidth}{!}{
\begin{tabular}{c|cc|ccc|ccc|c|c|c}
\toprule
\toprule
\multirow{2}{*}{\rule{0pt}{3ex}Model} & \multicolumn{2}{c|}{Acc} & \multicolumn{3}{c|}{S-BERT} & \multicolumn{3}{c|}{BERTScore} &
\multirow{2}{*}{\rule{0pt}{3ex}F1} &
\multirow{2}{*}{\rule{0pt}{4ex}\makecell{LLM as\\Judge}} &
\multirow{2}{*}{\rule{0pt}{3ex}Average}  \\\cmidrule(lr){2-3}\cmidrule(lr){4-6}\cmidrule(lr){7-9}
& P2 & P3 & P1 & P2 & P3 & P1 & P2 & P3 & & & \\
\midrule
Qwen2.5-VL-7B  & \underline{53.2} & 44.2 & 66.7 & 50.4 & 63.7 & 68.1 & 60.7 & 66.5 &\textbf{22.3} & 54.3 & 55.0 \\
InternVL3.5-8B  & \textbf{57.5} & 43.5 & 65.5 & 53.7 & 64.5 & 67.7 & 61.1 & 68.0 & \underline{19.5} & 55.5 & 55.6 \\
MiMo-V2-Flash-15B & 40.0 & 47.7 & 42.0 & 64.0 & 45.4 & 57.8 & 67.0 & 61.4 & 16.1 & 42.6 & 48.4 \\
EMIT-8B & 36.4 & {48.5} & \underline{69.3} & \underline{60.0} & \underline{70.2} & \underline{69.2} & \underline{67.1} & \textbf{70.0} & 10.5 & \underline{60.2} & \underline{56.1}\\
AD-Copilot-7B  & {52.5} & \textbf{53.0} & \textbf{74.5} & \textbf{70.0} & \textbf{71.2} & \textbf{70.8} & \textbf{70.1} & \underline{69.5} & 18.8 & \textbf{62.7} & \textbf{61.3} \\
\bottomrule
\bottomrule
\end{tabular}
}
\end{minipage}%
\hfill
\begin{minipage}[ht]{0.45\textwidth}
\centering
\caption{Comparison on the OmniDiff-Real Benchmark.}
\label{tab:omnidiff-real}
\setlength\tabcolsep{6pt}
\resizebox{\linewidth}{!}{
\begin{tabular}{l|cccc}
\toprule
\toprule
Method & BLEU-4 & METEOR & ROUGE-L & CIDEr \\
\midrule
GPT-4o                & 3.1 & \textbf{13.6} & 21.0 & 5.2 \\
LLaVA-OneVision-7B    & 0.2 &  4.1 & 13.6 & 1.7 \\
Qwen-2.5-VL-7B        & 3.8 &  9.5 & 19.8 & 6.2 \\
AD-Copilot-7B (Ours) & \textbf{5.2} & 12.6 & \textbf{21.5} & \textbf{7.4} \\
\bottomrule
\bottomrule
\end{tabular}
}
\end{minipage}

\end{table*}

% \begin{table}[htb]
% \centering
% \setlength\tabcolsep{3pt}
% \caption{Performance comparison on the UniPCB Benchmark.}
% \label{tab:unipcb}
% \resizebox{\linewidth}{!}{
% \begin{tabular}{c|cc|ccc|ccc|c|c|c}
% \toprule
% \toprule
% \multirow{2}{*}{\rule{0pt}{3ex}Model} & \multicolumn{2}{c|}{Acc} & \multicolumn{3}{c|}{S-BERT} & \multicolumn{3}{c|}{BERTScore} &
% \multirow{2}{*}{\rule{0pt}{3ex}F1} &
% \multirow{2}{*}{\rule{0pt}{4ex}\makecell{LLM as\\Judge}} &
% \multirow{2}{*}{\rule{0pt}{3ex}Average}  \\\cmidrule(lr){2-3}\cmidrule(lr){4-6}\cmidrule(lr){7-9}
% & P2 & P3 & P1 & P2 & P3 & P1 & P2 & P3 & & & \\
% \midrule
% Qwen2.5-VL-7B  & \underline{53.2} & 44.2 & 66.7 & 50.4 & 63.7 & 68.1 & 60.7 & 66.5 &\textbf{22.3} & 54.3 & 55.0 \\
% InternVL3.5-8B  & \textbf{57.5} & 43.5 & 65.5 & 53.7 & 64.5 & 67.7 & 61.1 & 68.0 & \underline{19.5} & 55.5 & 55.6 \\
% MiMo-V2-Flash-15B & 40.0 & 47.7 & 42.0 & 64.0 & 45.4 & 57.8 & 67.0 & 61.4 & 16.1 & 42.6 & 48.4 \\
% EMIT-8B & 36.4 & {48.5} & \underline{69.3} & \underline{60.0} & \underline{70.2} & \underline{69.2} & \underline{67.1} & \textbf{70.0} & 10.5 & \underline{60.2} & \underline{56.1}\\
% AD-Copilot-7B  & {52.5} & \textbf{53.0} & \textbf{74.5} & \textbf{70.0} & \textbf{71.2} & \textbf{70.8} & \textbf{70.1} & \underline{69.5} & 18.8 & \textbf{62.7} & \textbf{61.3} \\
% \bottomrule
% \bottomrule
% \end{tabular}
% }
% \end{table}

\paragraph{OmniDiff-Real benchmark}
We further evaluate AD-Copilot on OmniDiff-Real~\citep{liu2025omnidiff}, a general change captioning benchmark that measures the ability to describe visual differences between image pairs in everyday scenarios.
As shown in Tab.~\ref{tab:omnidiff-real}, general-purpose MLLMs achieve modest scores on this challenging benchmark.
AD-Copilot-7B, without any task-specific finetuning, clearly surpasses all evaluated MLLMs, demonstrating that the visual in-context comparison capability learned for IAD also transfers to non-industrial change captioning.

% \begin{table}[htb]
% \centering
% \setlength\tabcolsep{6pt}
% \caption{Performance comparison on the OmniDiff-Real benchmark.}
% \label{tab:omnidiff-real}
% \resizebox{\linewidth}{!}{
% \begin{tabular}{l|cccc}
% \toprule
% \toprule
% Method & BLEU-4 & METEOR & ROUGE-L & CIDEr \\
% \midrule
% GPT-4o                & 3.1 & \textbf{13.6} & 21.0 & 5.2 \\
% LLaVA-OneVision-7B    & 0.2 &  4.1 & 13.6 & 1.7 \\
% Qwen-2.5-VL-7B        & 3.8 &  9.5 & 19.8 & 6.2 \\
% AD-Copilot-7B (Ours) & \textbf{5.2} & 12.6 & \textbf{21.5} & \textbf{7.4} \\
% \bottomrule
% \bottomrule
% \end{tabular}
% }
% \end{table}

\subsection{Ablation Study}
\paragraph{Model Structure and Training Strategy Ablation} To validate the effectiveness of our training strategy and architecture, we conduct an ablation study on the MMAD benchmark, as shown in Tab.~\ref{ablation-study}.
The progressive inclusion of Stages~0--3 consistently improves the average accuracy, demonstrating the effectiveness of the Chat-AD training data and the soundness of our training recipe. 
Moreover, introducing the Comparison Encoder further boosts performance across all configurations, especially improving the F1 score of anomaly discrimination.
This indicates that AD-Copilot benefits from explicit multi-image comparison, enabling finer-grained difference perception between normal and anomalous regions.

\begin{figure}[htb]
\centering
\includegraphics[width=1.0\linewidth]{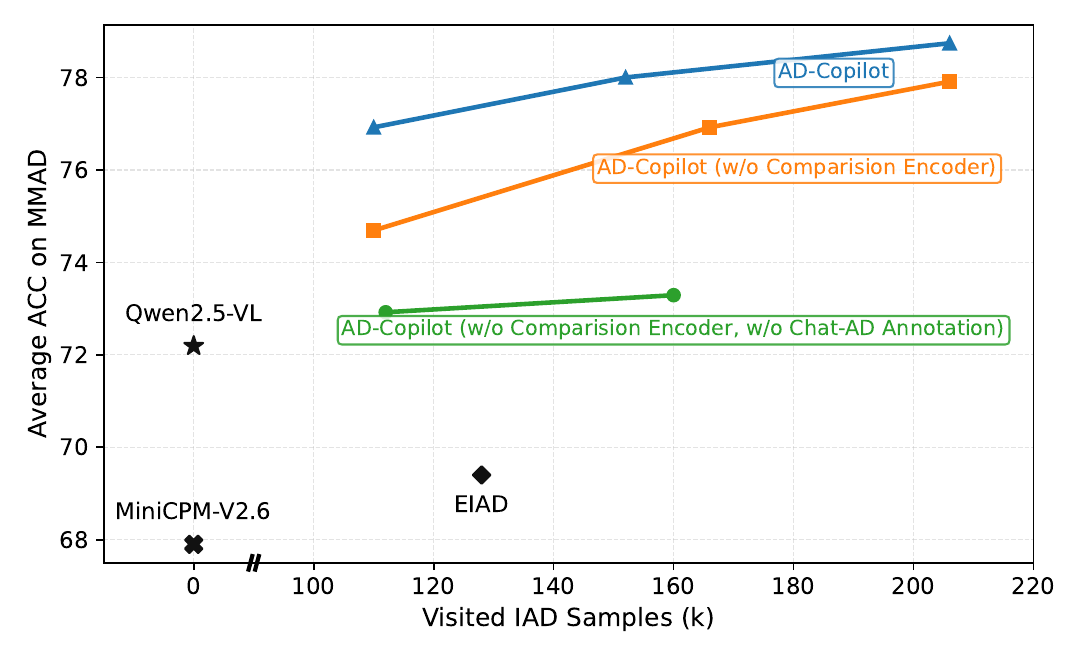} 
\vspace{-0.8em}
\caption{Data scaling behavior of AD-Copilot. Performance improves with more data, but removing the Comparison Encoder or using lower-quality annotations causes large drops.}\label{fig:scaling}
\end{figure}
\begin{table*}[htb]
\centering
\setlength\tabcolsep{3pt}
\caption{Ablation study of reward functions in the RL stage.}
\label{tab:reward-ablation}
% \resizebox{\linewidth}{!}
{
\begin{tabular}{c|c|c|cccc|cc|c}
\toprule\toprule
                       &                         & Anomaly        & \multicolumn{4}{c|}{Defect}                             & \multicolumn{2}{c|}{Object} &                           \\\cmidrule(r){3-9}
\multirow{-2}{*}{Model} & \multirow{-2}{*}{Scale} & Discrimination & Classification & Localization & Description & Analysis & Classification  & Analysis & \multirow{-2}{*}{Average} \\\midrule
\rowcolor{mygray}AD-Copilot-Thinking & 7B & 
\textbf{73.95} & \textbf{74.29} & \textbf{76.40} & \textbf{84.92} & 86.93 & 91.86 & \textbf{87.67} & \textbf{82.29} \\
w/o bbox reward & 7B & 
65.87 & 71.44 & 71.37 & 84.46 & \textbf{87.82} & 90.14 & 87.05 & 79.74 \\
w/o choice reward & 7B & 
65.24 & 65.81 & 62.47 & 77.67 & 84.63 & 89.30 & 82.27 & 75.34 \\
w/o RL & 7B & 
73.63 & 67.12 & 64.89 & 79.58 & 86.15 & \textbf{92.10} & 87.70 & 78.74 \\
\bottomrule\bottomrule
\end{tabular}
}
\end{table*}

\begin{table}[htb]
\centering
\setlength\tabcolsep{5pt}
\caption{Ablation study of AD-Copilot on the MMAD benchmark.}
\label{ablation-study}
% \resizebox{\linewidth}{!}
{
\begin{tabular}{ccccc|cc}
\toprule\toprule
\makecell[c]{Comparison\\Encoder} & Stage 0 & Stage 1 & Stage 2 & Stage 3 & ACC & F1 \\
\midrule
 &  &  &  &  & 72.19 & 72.49 \\
 &  &  & \checkmark  &  & 77.28 & 65.65 \\
\checkmark  &  &  & \checkmark  &  & 76.92 & 73.27 \\
 & \checkmark  &\checkmark &\checkmark &  & 77.91 & 73.60 \\
\checkmark  &\checkmark &\checkmark &\checkmark &  & 78.74 & \textbf{75.14} \\
 &\checkmark &\checkmark &\checkmark &\checkmark & \underline{80.46} & 71.73 \\
\checkmark &\checkmark &\checkmark &\checkmark &\checkmark & \textbf{82.29} & \underline{74.23} \\
\bottomrule\bottomrule
\end{tabular}
}
\end{table}

\paragraph{Data Scaling Behavior}
We further study the effect of data scaling in Fig.~\ref{fig:scaling}. Performance steadily improves with more training data, but removing the Comparison Encoder or using Chat-AD images with only sketchy annotations (instead of our carefully curated data) causes substantial performance drops. This demonstrates that scaling data alone is insufficient for IAD; both knowledge injection through high-quality data curation and perceptual enhancement through the Comparison Encoder are critical.

% \paragraph{Comparison Token Design}
% The number of comparisons is determined by the diagonal positional embedding design, which requires it to equal the square root of the maximum image token count ($\sqrt{10000} = 100$). We explore an alternative design using the positional encoding of the first 200 image tokens with 200 comparison tokens, which results in a performance drop from 78.74 to 77.28 in average accuracy. This indicates that our diagonal encoding scheme better captures the spatial locations of differences across the full image extent.

\paragraph{Reward Function Ablation}
As discussed in Sec.~\ref{subsec:recipe}, we employ two types of verifiable questions during the reinforcement learning stage and accordingly design two reward functions. To understand their contributions, we conduct ablation studies, as shown in Tab.~\ref{tab:reward-ablation}. Although the bbox localization task and the multiple-choice task differ significantly in form, the results clearly show that removing the bbox reward leads to substantial drops in both anomaly discrimination and defect localization. This indicates that bbox-based localization questions effectively strengthen the model's understanding of anomalies and enhance both perception and localization capabilities, which in turn leads to improved performance on choice-based tasks. When the choice reward is removed, performance drops even below the setting without reinforcement learning. This occurs because training only with bbox-style questions creates an optimization bias that pulls the model away from multiple-choice reasoning. Therefore, both reward functions play complementary roles for AD-Copilot's SOTA performance.

\subsection{Additional Analyses}

\paragraph{Effect of the Comparison Encoder}

\begin{figure}[htb]
\centering
\includegraphics[width=1.0\linewidth]{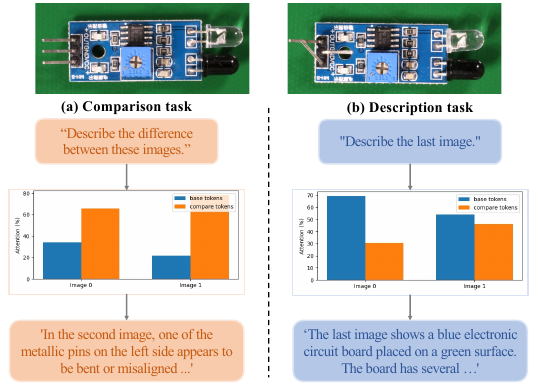} 
\caption{Visualization of the attention ratio between base tokens and comparison tokens across different tasks.}\label{attention}
\end{figure}

To explore how the Comparison Encoder improves model performance, we visualize the attention ratio between base tokens and comparison tokens across different tasks in Fig.~\ref{attention}. When the task involves image comparison (e.g., anomaly discrimination, defect description), the attention weight on comparison tokens increases significantly, indicating their active role in encoding inter-image differences and enhancing IAD performance. In contrast, when describing a single industrial image, attention focuses more on base tokens, suggesting that comparison tokens do not interfere with basic perception.

To further validate the generality of the Comparison Encoder, we directly plug the trained Comparison Encoder from AD-Copilot into Qwen2.5-VL-3B \textbf{without any additional finetuning}. Given the shared vision encoder architecture, comparison tokens are effectively transferred via a linear embedding-space alignment. As shown in Tab.~\ref{tab:3B}, the Comparison Encoder provides clear performance gains on the anomaly discrimination task, particularly in recall (+4.5 points) and F1 score (+2.4 points), confirming its strong cross-model generalization capability.

\begin{table}[htb]
\centering
\setlength\tabcolsep{5pt}
\caption{Transfer of the Comparison Encoder to Qwen2.5-VL-3B on the Anomaly Discrimination task of MMAD (without finetuning).}\label{tab:3B}
% \resizebox{\linewidth}{!}
{
\begin{tabular}{c|c|c|c|c}
\toprule
\toprule
Model & Accuracy & Recall & Precision & F1 \\
\midrule
Qwen2.5-VL-3B & 61.51 & 53.19 & \textbf{74.87} & 60.88\\
+ Comparison Encoder & \textbf{62.69} & \textbf{57.69} & 73.89 & \textbf{63.27} \\
\bottomrule
\bottomrule
\end{tabular}
}
\end{table}

\paragraph{Multi-shot Generalization}
The Comparison Encoder naturally extends to multi-shot settings by comparing each image with its neighbors, thereby enhancing contextual perception. Although AD-Copilot is trained exclusively in a 1-shot setting, it generalizes well to multi-shot scenarios and maintains leading performance, as shown in Tab.~\ref{tab:muti-shot}. Specifically, AD-Copilot achieves consistent improvements over the Qwen2.5-VL baseline across all shot settings (0, 2, 4, 8), while some other IAD-specific methods have been reported to degrade in multi-shot settings.

\begin{table}[htb]
\centering
\setlength\tabcolsep{8pt}
\caption{Multi-shot experiment on MMAD (Average ACC).}
\label{tab:muti-shot}
% \resizebox{\linewidth}{!}
{
\begin{tabular}{c|c|c|c|c}
\toprule
\toprule
Model & 0-shot & 2-shot & 4-shot & 8-shot  \\
\midrule
Qwen2.5-VL-7B & 68.91  & 72.20 & 72.29 & 71.63\\
AD-Copilot-7B & \textbf{73.35} & \textbf{77.53} & \textbf{77.42} & \textbf{76.67} \\
\bottomrule
\bottomrule
\end{tabular}
}
\end{table}

%%%%%%%%%%%%%%%%%%%%%%%%%%%%%%
\begin{figure*}[htb]
\centering
\includegraphics[width=1.0\linewidth]{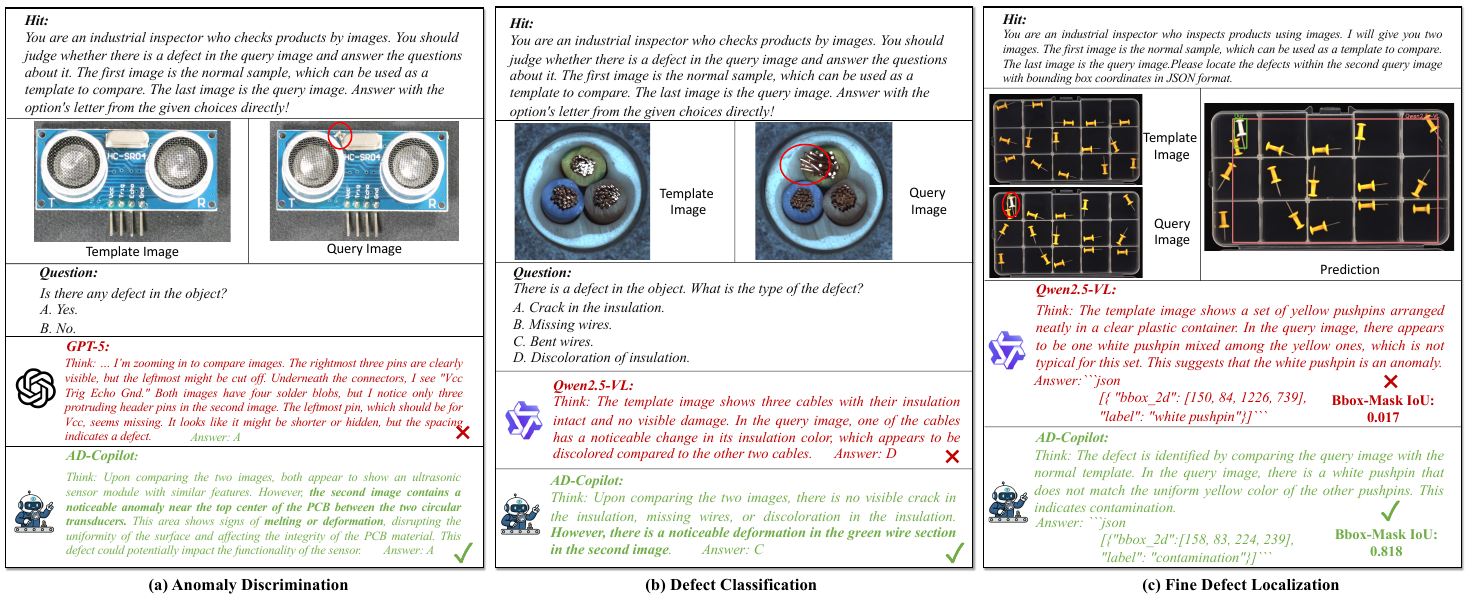} 
\vspace{-0.8em}
\caption{Qualitative evaluation of AD-Copilot compared with other models across various IAD tasks.}\label{qualitative-evaluation}
\end{figure*}
%%%%%%%%%%%%%%%%%%%%%%%%%%%

\paragraph{Incorporating Domain Knowledge}
In previous works, domain knowledge in text form is often used as an additional prior to boost performance under default settings. However, the gains are heavily related to the text context, and this kind of prior may not be available in real industrial applications. Therefore, we report the effect of domain knowledge as supplementary results. As shown in Tab.~\ref{tab:domain-knowledge}, incorporating domain knowledge during inference results in a slight improvement in both accuracy and F1 score. Furthermore, applying domain knowledge consistently in both training and inference brings a more significant gain, demonstrating that knowledge-aware learning can further enhance model generalization.

\begin{table}[ht]
\centering
\setlength\tabcolsep{5pt}
\caption{Effect of incorporating domain knowledge.}
\label{tab:domain-knowledge}
% \resizebox{\linewidth}{!}
{
\begin{tabular}{l|cc}
\toprule
\toprule
Model Setting & ACC & F1 \\
\midrule
AD-Copilot-Thinking & 82.29 & 74.23 \\
+ Domain Knowledge (Inference) & \underline{82.84} & \underline{74.59} \\
+ Domain Knowledge (Training \& Inference) & \textbf{83.15} & \textbf{77.60} \\
\bottomrule
\bottomrule
\end{tabular}
}
\end{table}

\paragraph{Effect of Template Image Selection}
As template images provide critical reference information for visual in-context comparison in MMAD experiments, their selection can significantly influence the results. As shown in Tab.~\ref{tab:template-image}, removing the template (0-shot) causes a substantial performance drop, highlighting the importance of template images in guiding anomaly understanding. Conversely, substituting the default template with the one most similar to the query image (1-shot+) leads to further improvement. This confirms that AD-Copilot effectively utilizes template images to capture normal patterns and perform precise visual comparison---the core capability our framework is designed to enhance.

\begin{table}[htb]
\centering
\setlength\tabcolsep{5pt}
\caption{Effect of template image settings on model performance.}
\label{tab:template-image}
\begin{tabular}{l|cc}
\toprule
\toprule
Setting & ACC & F1 \\
\midrule
0-shot (No Template) & 79.15 & 36.41 \\
1-shot (Default, Random Template) & \underline{82.29} & \underline{74.23} \\
1-shot+ (Retrieved Template) & \textbf{83.37} & \textbf{74.85} \\
\bottomrule
\bottomrule
\end{tabular}
\end{table}

\paragraph{Qualitative Analysis}
We present several qualitative examples in Fig.~\ref{qualitative-evaluation}, comparing AD-Copilot with other MLLMs. In the first example, the latest GPT-5 model exhibits hallucinated reasoning (assuming the pins are cut off) rather than identifying the actual defect (solder splash). In contrast, AD-Copilot not only gives the correct answer but also precisely identifies the defect's shape and position in its reasoning process. 
In the second example, Qwen2.5-VL fails to recognize the correct defect type, while AD-Copilot accurately distinguishes it after domain-specific training. 
The third example demonstrates the BBox-based localization task, where AD-Copilot correctly localizes the anomaly, whereas Qwen2.5-VL, despite having found the anomaly, fails to localize it precisely.

%% ==================== CONCLUSION ====================
\section{Conclusion}\label{sec:conclusion}

We introduce \textbf{AD-Copilot}, a vision-language assistant for industrial anomaly detection built around the paradigm of \textit{visual in-context comparison}.
We identify the semantic comparison bottleneck as the fundamental limitation preventing existing MLLMs from leveraging template images for fine-grained anomaly detection.
To address this, we construct \textbf{Chat-AD}, a large-scale comparison-centric dataset with over 620,000 samples across 327 industrial categories; design the \textbf{Comparison Encoder}, a plug-and-play module that generates comparison tokens via cross-attention to enable visual-level comparison; and develop a \textbf{multi-stage training curriculum} that progressively builds comparison capability from general visual differences to industrial anomaly detection and reasoning.
We further establish \textbf{MMAD-BBox} with the BBox-Mask IoU metric for standardized evaluation of anomaly localization.
Extensive experiments demonstrate that AD-Copilot achieves state-of-the-art results across many benchmarks, surpassing all baselines and even outperforming human experts in several tasks.
Together, these contributions represent a significant step toward integrating MLLMs as reliable assistants in real-world industrial inspection.

%% ==================== BIBLIOGRAPHY ====================
\bibliographystyle{IEEEtran}
\bibliography{main}

\end{document}